\documentclass[runningheads,a4paper]{llncs}

\usepackage[utf8]{inputenc} 

\usepackage{amsmath}
\usepackage{amssymb}
\usepackage{graphicx}
\usepackage{color, colortbl} 
\usepackage{graphicx} 
\usepackage{array}

\usepackage{url}
\newcommand{\keywords}[1]{\par\addvspace\baselineskip
\noindent\keywordname\enspace\ignorespaces#1}

\usepackage[colorinlistoftodos]{todonotes}
\usepackage[colorlinks]{hyperref}
\usepackage[]{booktabs}

\usepackage{mdframed}

\usepackage{etoolbox}

\definecolor{Gray}{gray}{0.8}
\newcolumntype{g}{>{\columncolor{Gray}}c}
\newcolumntype{k}{>{\centering\arraybackslash}p{13mm}}
\newcolumntype{f}{>{\centering\arraybackslash}p{12mm}}
\newcolumntype{h}{>{\centering\arraybackslash}p{5mm}}
\newcolumntype{j}{>{\centering\arraybackslash}p{8mm}}

\begin{document}

\mainmatter  

\title{Sign Languague Recognition without frame-sequencing constraints: A proof of concept on the Argentinian Sign Language}

\titlerunning{A General Model for Sign Language Recognition}

\author{Franco Ronchetti* \inst{1} \and Facundo Quiroga* \inst{1} \and César Estrebou \inst{1} \and Laura Lanzarini \inst{1} \and Alejandro Rosete \inst{2}}

\authorrunning{F. Ronchetti et al.}

\def\aa{\}}
\def\ab{\{}

\institute{
Instituto de Investigación en Informática LIDI, Facultad de informática,\\
  Universidad Nacional de La Plata
 \\ \email{\ab fronchetti,fquiroga,cesarest,laural\aa @lidi.info.unlp.edu.ar}
\and
  Instituto Superior Politécnico Jose Antonio Echeverría
 \\ \email{\ab rosete\aa @ceis.cujae.edu.cu} \footnotetext{* Contributed equally }
}

\toctitle{}
\tocauthor{}
\maketitle

\begin{abstract}

Automatic sign language recognition (SLR) is an important topic within the areas of human-computer interaction and machine learning. On the one hand, it poses a complex challenge that requires the intervention of various knowledge areas, such as video processing, image processing, intelligent systems and linguistics. On the other hand, robust recognition of sign language could assist in the translation process and the integration of hearing-impaired people, as well as the teaching of sign language for the hearing population.

SLR systems usually employ Hidden Markov Models, Dynamic Time Warping or similar models to recognize signs.  Such techniques exploit the sequential ordering of frames to reduce the number of hypothesis. This paper presents a general probabilistic model for sign classification that combines sub-classifiers based on different types of features such as position, movement and handshape. The model employs a bag-of-words approach in all classification steps, to explore the hypothesis that ordering is not essential for recognition. The proposed model achieved an accuracy rate of 97\% on an Argentinian Sign Language dataset containing 64 classes of signs and 3200 samples, providing some evidence that indeed recognition without ordering is possible.

\end{abstract}

\section{Introduction}

\subsection{Background}

Automatic sign recognition is a complex, multidisciplinary problem that has not been fully solved. While recently there has been some lateral progress through gesture recognition, driven mainly by the development of new technologies, there is still a long road ahead before accurate and robust applications are developed that allow translating and interpreting the signs performed by an interpreter\cite{Cooper2011a}. The complex nature of signs draws effort in various research areas such as human-computer interaction, computer vision, movement analysis, machine learning and pattern recognition.

The full task of recognizing a sign language involves a multi-step process \cite{Cooper2011a}, which can be simplified as:
\begin{enumerate}
  \item Tracking the hands of the interpreter
  \item Segmenting the hands and creating a model of its shape
  \item Recognizing the shapes of the hands
  \item Recognizing the sign as a syntactic entity.
  \item Assigning semantics to a sequence of signs.
  \item Translating the semantics of the signs to the written language
\end{enumerate}

While these tasks can provide feedback to each other, they can be carried out mostly independently, and in different ways. For example, there are several approaches for tracking hand movements: some use 3D systems \cite{moreira2014recognizing,PugeaultBowden2011b}, such as MS Kinect, and others simply use a 2D image from an RGB camera \cite{Cooper2011a,von2008recent}. Most older systems employed movement sensors such as special gloves, accelerometers, etc, but recent approaches generally focus on video, such as the one presented here.

There are numerous publications dealing with the automated recognition of sign languages, a field that started mostly in the 90s. Von Agris \cite{von2008recent} and Cooper \cite{Cooper2011a} both present a general view of the state of the art in sign language recognition.

Sign language recognition employs different types of features, usually classified as manual and non-manual.

Non-manual features such as pose, lip-reading or face expressions are sometimes included to enhance the recognition process, since some signs cannot be differentiated from manual information only \cite{von2008recent}. In this regard, the tracking of the head is mostly solved \cite{viola2004robust}, but its segmentation with respect to an arbitrary background or in the presence of hand-head occlusions is still an unsolved problem, as is the robust recognition of such non-manual features \cite{von2008recent}. Manual information, on the other hand, generally conveys most of the information in a sign.

For tracking and segmentation of the hands, there is much interest in creating skin color models to detect and track the hands of an interpreter on a video \cite{Roussos2010a}, and adding the possibility of segmenting the hands \cite{Cooper2012}, even in the presence of hand-hand occlusions \cite{zieren2005robust}.

The handshape information of a sign is composed by a sequence of hand poses. After segmenting the hand, it must represented in a convenient way for handshape recognition. However, turning a hand pose into another requires a non-rigid transformation of the hand, which must be also modeled, and capturing non-rigid 3D transformations with occlusions using a 2D RGB camera is a hard task. While the best possible output from this step would be a full 3D model of the hand, this is generally hard to do without multiple cameras, special sensors or markers \cite{PugeaultBowden2011b}. In most cases the handshape is instead represented as a combination of more abstract features based on geometric or morphologic properties of its shape or texture \cite{von2008recent}.

Some researchers focus on fingerspelling \cite{PugeaultBowden2011b}, which is essentially a static handshape recognition task. While some signs do indeed present a static handshape in one or both hands, and no movement, most involve many handshapes and their transitions (i.e., non-rigid body transformations of the hand), or rigid body transformations of a single handshape (i.e., rotation and traslation), and a certain movement of the hands. To deal with these dynamic signs, SLR systems are based usually on Hidden Markov Models (HMMs), Dynamic Time Warping (DTW) or similar models, whether to recognize segmented signs or a continuous stream \cite{von2008recent,Cooper2011a}. These techniques attempt to model the sequence of positions and handshapes, therefore exploiting the sequential ordering of frames to reduce the number of hypothesis to test.

Finally, SLR techniques have traditionally employed sign-level models to recognize sign language. Recently, there have been many efforts to move from sign-level models to sub-unit level models, analogously to the transition from word-level to phoneme-level models in speech recognition \cite{Cooper2012}. The main problem found in these attempts has been the difficulty to find a useful definition of sub-units of a sign. Unlike speech, signing is very much multimodal and there is little standardization of both languages and specification languages, therefore a promising approach is to infer such sub-units from training data \cite{von2008recent}.

\subsection{Presented work}

This paper presents a general classification model for sign language recognition that focuses on step 4, that is, the recognition of signs as a syntactic entity on the sign-level (i.e., a correspondence between a video containing a sign and a word). The model is composed of a set of subclassifiers, each employing a bag-of-words approach that ignores sequence information. This setup allows us to explore the hypothesis that a classifier can still achieve high recognition accuracies under such a constraint.


While models that exploit the ordering of frames such as HMMs should theoretically achieve greater recognition accuracy, such constraint complicates the inference of sub-units, since by definition they must too be ordered. The bag of words approach can ease the task of determining sub-units, since their sequence or transitions do not need to be specified or learned.

To test the model, we performed experiments on the LSA64 dataset, which consists of 64 signs of the Argentinian Sign Language (LSA) and was recorded with normal RGB cameras.

The document is organized as follows. Section 2 describes the LSA64 dataset, the image processing and feature extraction. Section 3 defines the classification model. Section 4 details the experiments carried out, and finally Section 5 presents the general conclusions.

\section{Dataset and Features}

  \subsection{Argentinian Sign Language Dataset (LSA64)}

The sign dataset for the Argentinian Sign Language \footnote{More information about this dataset can be found at \url{http://facundoq.github.io/unlp/lsa64/.}}, includes 3200 videos where each of the 10 non-expert subjects performed 5 repetitions of 64 different types of signs.

\begin{figure}
\centering
\includegraphics[width=0.24\textwidth]{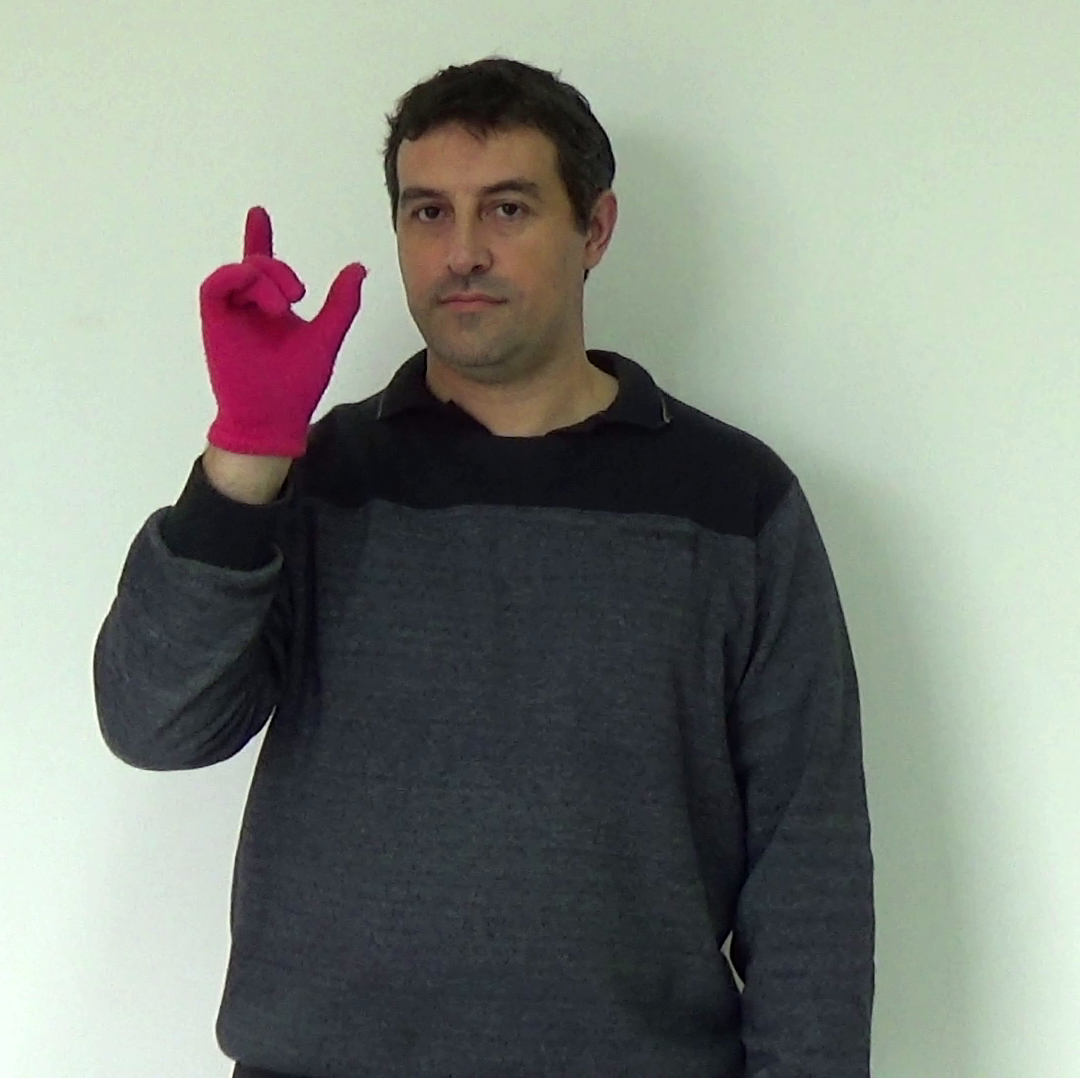}
\includegraphics[width=0.24\textwidth]{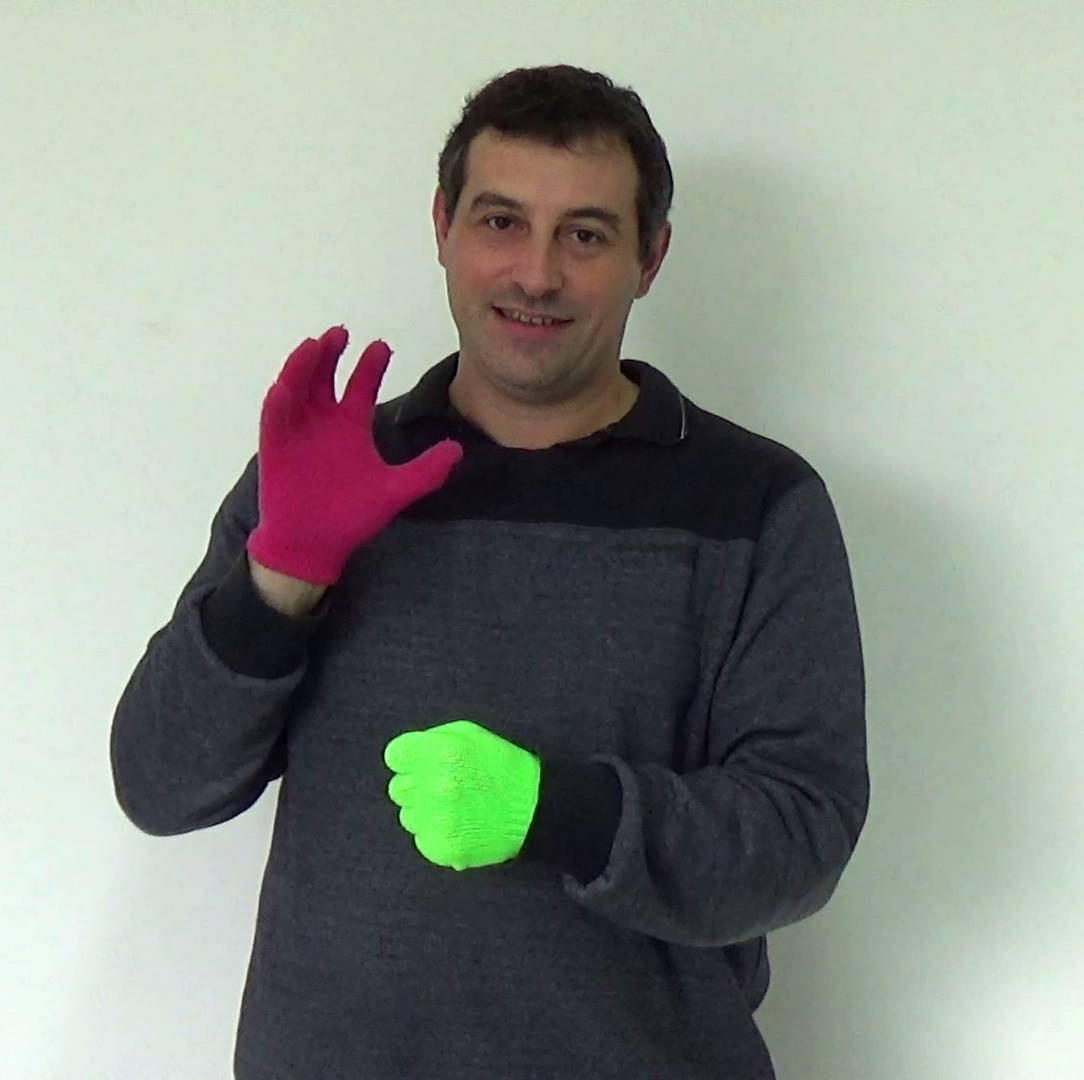}
\includegraphics[width=0.24\textwidth]{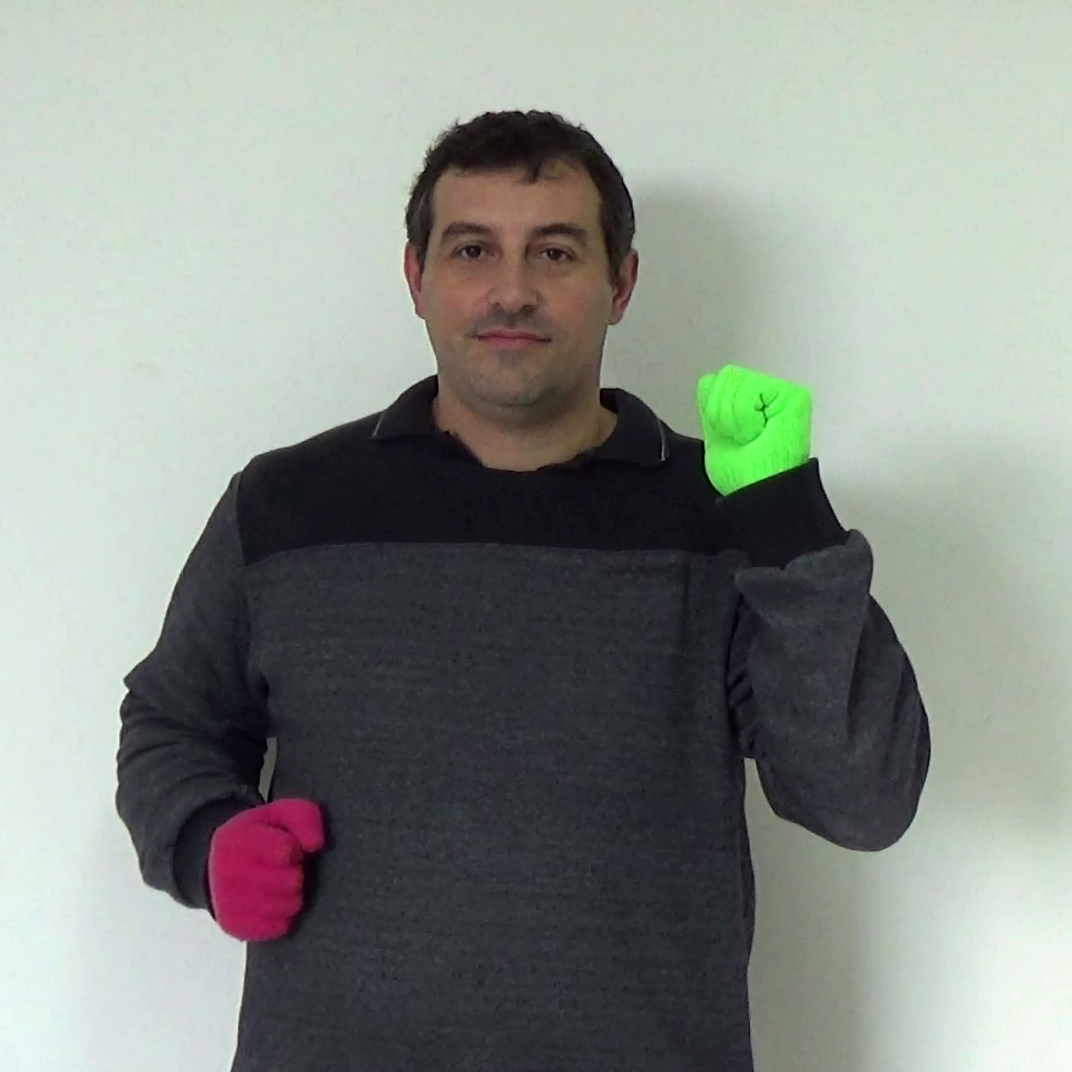}
\includegraphics[width=0.24\textwidth]{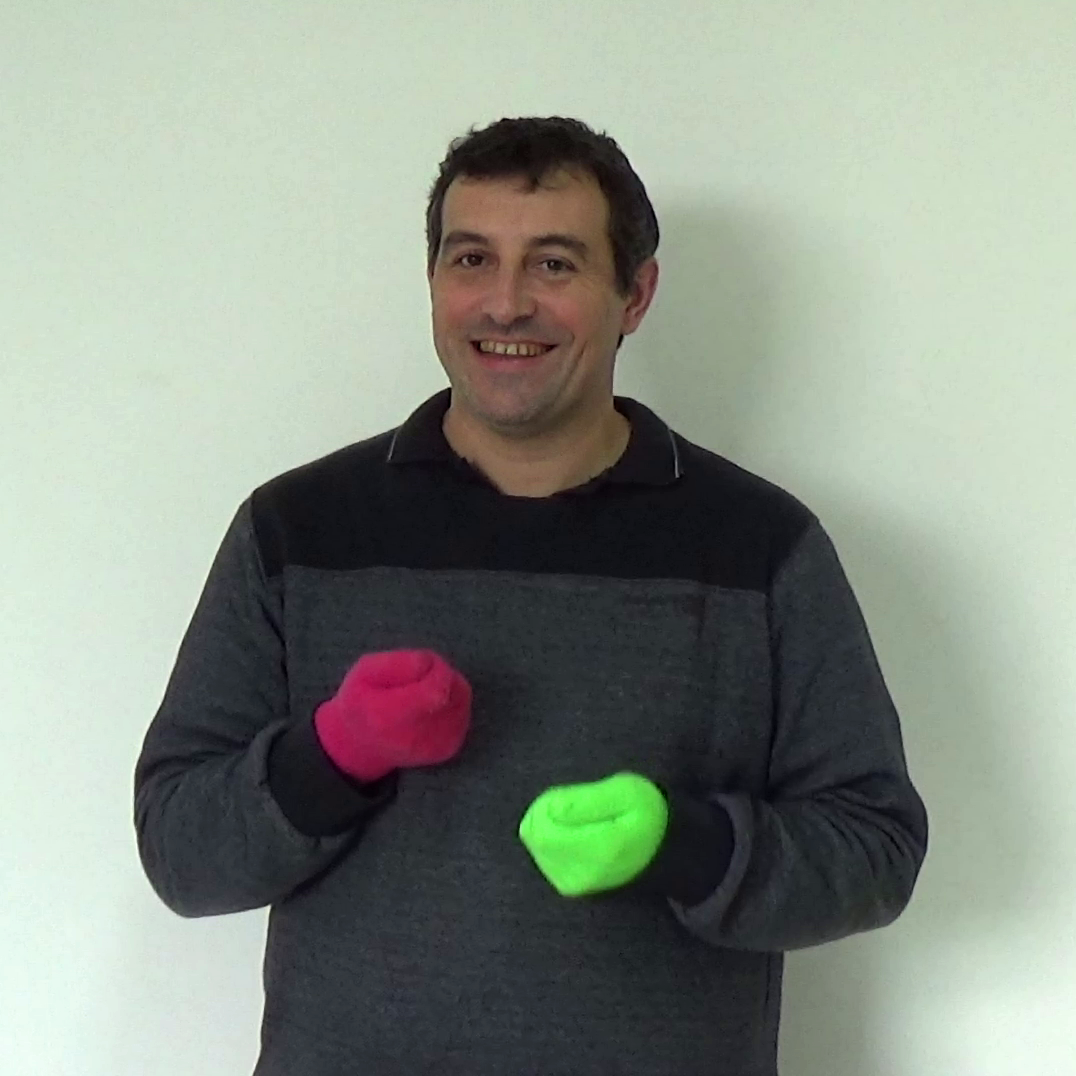}
\caption{Snapshots of the performance of four different signs of the LSA64 database.}
\label{fig:lsa64}
\end{figure}

To simplify the problem of hand segmentation within an image, subjects wore fluorescent-colored gloves, as can be seen in Figure \ref{fig:lsa64}. The glove substantially simplifies the problem of recognizing the position of the hand and performing its segmentation by removing all issues associated to skin color variations and hand-head occlusions, while fully retaining the difficulty of recognizing the handshape.

The dataset contains 22 two-handed signs and 42 one-handed ones, which were selected among the most commonly used ones in the LSA lexicon.

\vspace{-30px}

  \subsection{Preprocessing and Features}
  The pre-processing and feature extraction activities carried out with the database's videos consist of extracting hand and head positions, along with images of the hand, segmented.

The detection of the hand, and generation of features based on the hand images for each video frame is based on the work of \cite{Ronchetti2016}. Additionally, the head of the subjects is tracked via the Viola-Jones's face detector \cite{viola2004robust}. The 2D position of each hand is then transformed to be relative to that of the head. The positions are normalized by dividing by the arm's length of the subject, measured in centimeters/pixels. In this way, the transformed positions represent displacements from the head, in units of centimeters.

The result of this process is a sequence of frame informations, in which for each frame we calculate the normalized position of both hands, and we extract an image of each hand with the background segmented.

\section{Classification Model}

\begin{figure}
	\centering
	\includegraphics[width=0.98\textwidth]{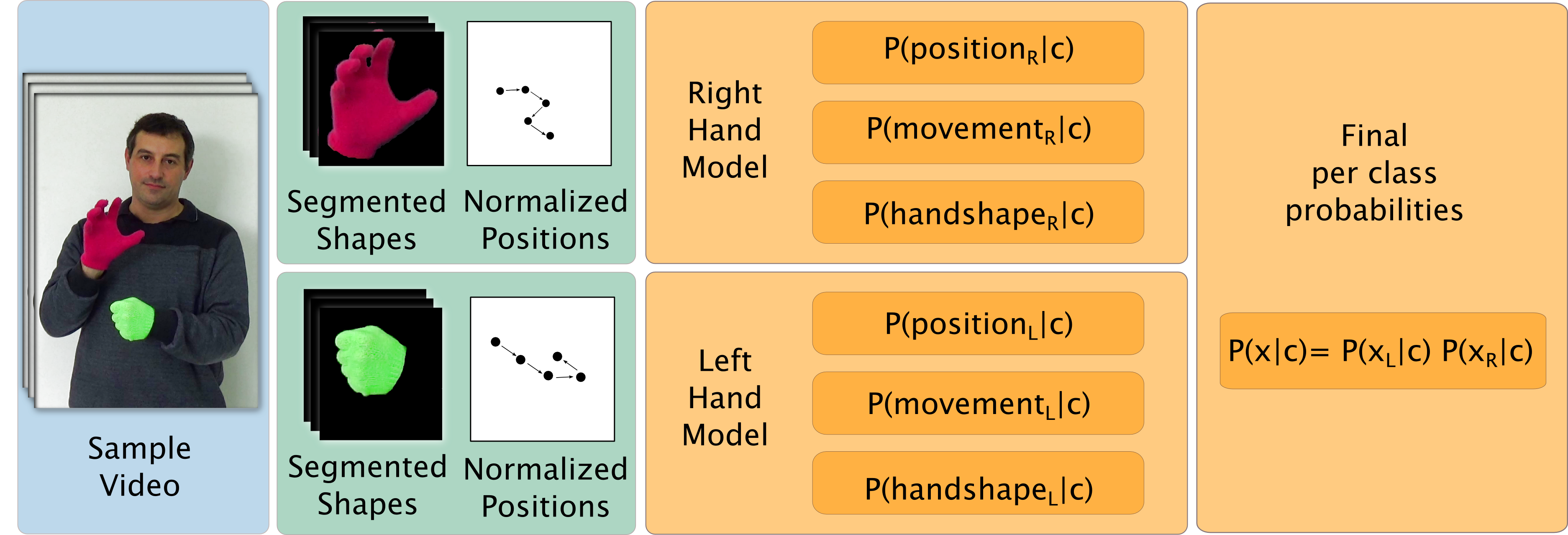}
	\caption{Diagram of the model.}
	\label{fig:model}
\end{figure}

The model combines the output of two subclassifiers, one for each hand. The subclassifier for each hand combines as well the output of three other subclassifiers that each use position, movement and handshape information (Figure \ref{fig:model}).

From the sample video, the sequence of segmented hand images and hand positions is extracted. For each hand, the position info is fed to the position and movement subclassifiers, while the segmented hand images are used as input for the handshape subclassifier. The right hand and left hand subclassifiers each output probabilities for each class, and the final output combines those per class.

Since the proposed model classifies each hand separately, we define the probability of a sample sign $x$ given a class of sign $c$ as:

\begin{equation}
  \label{eq:model1}
  P(x|c)=P(x^l|c) P(x^r|c)
\end{equation}

\noindent where $x^l$ and $x^r$ is the sample sign information in the left and right hand respectively. Being able to split the probabilities in this way depends on the (naive) assumption that the synchronization between the hands is not important for the recognition of the sign, or at least that such information is not crucial for recognition.

The classifier for a single hand depends on several subclassifiers that focus on specific parts of the sign.
Since the same type of subclassifiers are employed for each hand, in the following we assume that $h$ can be either $l$ or $r$, for \textit{left} and \textit{right}.

The three subclassifiers for a hand $h$ use the information of the sequence of positions of the hand of the sample, $x_p^h$, the movement it performs, $x_m^h$, and the sequence of handshapes it goes through, $x_s^h$. Therefore the probability for a hand can be written as:

\begin{equation}
  \label{eq:pxh}
  P(x^h|c)=P(x_p^h|c) P(x_m^h|c) P(x_s^h|c)
\end{equation}

As before, Equation \ref{eq:pxh} assumes independence between $x_p^h$, $x_m^h$ and $x_s^h$, that is, that the position of the hand in no way restricts the types of possible movements or configurations, etc. Since signers are usually restricted to move their hands inside an imaginary square centered at their torso, this assumption holds in most practical cases.


Hence, to classify a sample sign $x$ we pick the class of sign $c$ with the maximum probability $P(x|c)$, that expands into:

\begin{equation}
  P(x|c)=P(x^l|c) P(x^r|c) = P(x_p^l|c) P(x_m^l|c) P(x_s^l|c) P(x_p^r|c) P(x_m^r|c) P(x_s^r|c)
\end{equation}

Table \ref{tab:symbols} describes symbols of the notation. In the following subsections, we describe how we calculate the probability for each subclassifier, and how we extend this model to deal with the absence of a hand and signs in which a hand does not move.

\newcolumntype{s}{>{\centering\arraybackslash}m{7mm}}
\newcolumntype{e}{>{\centering\arraybackslash}m{20mm}}
\begin{table}
\centering
\caption{Notation reference. The variable $x$ refers always to sample information. Subscripts indicate type of information. Superscripts indicate the hand. Variable $a$ refers to a parameter of the model.}
\begin{tabular}{sesesese}
  \toprule
   $c$ & Class & $x$ & Sample & $h$ & Hand $h$ (generic) & $(\cdot)^h$ & Hand $h$ info \\
	 \midrule
   $(\cdot)^l$ & Left hand info & $(\cdot)^r$ & Right hand info  & $x_p^h$ & Position  & $x_s^h$ & Handshape  \\
	 \midrule
  $x_m^h$ & Movement & $x_{tm}^h$ & Trajectory & $x_{am}^h$ & Amount of movement & $x_a^h$ & Absence of hand $h$ in testing \\
	\midrule
    $a_c^h$ & Hand $h$ is not used in class $c$  & $a_{c,m}^h$ & Absence of movement in training &  &  &  &  \\
 \bottomrule
\end{tabular}
\label{tab:symbols}
\end{table}

\subsection{Position-based probability}
\label{sec:position}

From the set of absolute positions a hand goes through in the execution of the sign, we only employ the first ($fp$) and last ($lp$).  We performed a 2D Kolmogorov-Smirnov normality test on the first and last positions and found that with a confidence level of 95\% there is enough evidence to reject the hypothesis of normality in 30\% of the classes. However, when fitting the positions with a Gaussian Mixture Model we found that for 78.5\% of the models a single component provided the best Bayesian Information Criteria score, and that performance on the test set was lower than with a single gaussian, possibly due to overfitting avoidance. Therefore, we chose to model the positions for each class using a single 2D normal distribution.

From the training set data we calculate the means $\mu_{fp,c}$ and $\mu_{lp,c}$ and covariances $\Sigma_{fp,c}$ and $\Sigma_{lp,c}$ of the first and last positions, for each class $c$. The probability for a new sample with position information $x^p$ given class $c$ is computed as:

\begin{equation}
  \label{eq:model_position}
  P(x_p^h|c) = g_{fp,c}(x_p^h)  g_{lp,c}(x_p^h)
\end{equation}

\noindent where $g_{fp,c}$ is a 2D gaussian probability density function with mean $\mu_{fp,c}$ and covariance $\Sigma_{fp,c}$. The gaussian $g_{lp,c}$ is defined analogously for the last position $lp$.

\subsection{Movement-based probability}

We consider two factors for the movement-based probability, based on trajectory ($x_{tm}$) and amount of movement ($x_{am}$) information, so that:

\begin{equation}
  \label{eq:model_movement}
  P(x_m^h|c) = P(x_{tm}^h|c)  P(x_{am}^h|c)
\end{equation}

\subsubsection{Amount of movement}
\label{sec:amountmovement}
We calculate the amount of movement for a hand $h$ and a sign $c$ as the maximum distance between two positions of the hand. In the training phase, we compute the mean amount of movement $\mu_{am,c}^h$ along with standard deviation $\sigma_{am,c}^h$. During testing, we penalize classes for which the movement of testing sample $x$ differs greatly from $\mu_{am,c}^h$, so that:

\begin{equation}
  \label{eq:model_movement_amount}
  P(x_{am}^h|c)=g_{am,h}(x_{am}^h)
\end{equation}

\noindent where $g_{am,h}$ is a 1D Gaussian probability density function with mean and standard deviation $\mu_{am,c}^h$ and $\sigma_{am,c}^h$.

\subsubsection{Trajectory}

We calculate the probability for the trajectory of the hand, $P(x_{tm}^h|c)$, adapting a classifier which was employed succesfully in \cite{Ronchetti2015} for action recognition. In this model, a movement of the hand is described as a discrete path in space, i.e., a list of positions in space. The model calculates the set of directions of that discrete path in space, which is formed by the set of (normalized) vector difference between hand positions in successive frames. By quantizing the possible directions, the classifier computes a distribution of directions of a sample movement, which is then compared to the distribution of directions of training samples to determine the probability of a sample for each class.

This subclassifier employs some sequence information, since to compute the directions of the sign, we need to compute the vector difference between positions of consecutive frames. However, after computing this difference, the order of the directions is irrelevant for the subclassifier. Moreover, if we consider each direction as an estimate of the instantaneous unit velocity of the hand at each frame, we can see that this scheme is simply a proxy for computing the velocities in each frame.

\subsubsection{Signs with no movement in one or both hands}

Signs in which a hand does not move present a problem. If a class $c$ has very little or no movement, then the trajectory probability $P(x_{tm}^h|c)$ is not useful and will probably penalize classes randomly. To avoid this situation, we employ again the per-class amount of movement means $\mu_{am,c}^h$ computed at training time, and we set a parameter:

\begin{equation}
a_{c,m}^h=\begin{cases}
1 & \mbox{if } \mu_{am,c}^h > 5cm \\
0 & \mbox{if } \mu_{am,c}^h \leq 5cm
\end{cases}
\end{equation}

\noindent where we determined the threshold as 5cm experimentally. We can use this parameter as an exponent for $P(x_{tm}^h|c)$ to neutralize that factor for classes with no or little movement by redefining Equation \ref{eq:model_movement} as:

\begin{equation}
  \label{eq:model_movement_no}
  P(x_m^h|c) = P(x_{tm}^h|c)^{a_{c,m}^h}  P(x_{am}^h|c)
\end{equation}

In this way, if a sign does not have movement in a hand, by setting $a_{c,m}^h=0$ the model can ignore the trajectory information.


\subsection{Handshape-based probability}

\label{sec:handshape}

To obtain the probabilities of each class for the sequence of handshapes of a sign $x$, we first calculate the probability for each handshape on each frame of a sign, based on the segmented 2D image of the hand, using the classifier described in \cite{Ronchetti2016}. Then, we use the probabilities for all frames to model the sequence of handshapes and transformations of the hand.

To calculate the handshape-based probability for the whole sign, we also employ a similar approach as in the trajectory-based subclassifier, but now instead of quantizing over directions, we quantize over the set of vectors that indicate the probability that the hand is in a given hand pose.


\subsection{One-handed signs}

\subsubsection{Ignoring the information of an unused hand}

Some classes of signs only use one hand. It is possible, however, that at recognition time the other hand is not off camera but present in the video with random positions, movements or handshapes. It is desirable not to consider the other hand's information for one-handed signs, since such random information could be interpreted by the classifier as a genuine attempt to perform part of a sign. To avoid those situations, we can turn Equation \ref{eq:model1} into:

\begin{equation}
  \label{eq:model2}
  P(x|c)=P(x^l|c)^{a_c^l} P(x^r|c)^{a_c^r}
\end{equation}

\noindent where $a_c^l$ and $a_c^r$ have value either $1$ or $0$. Setting $a_c^h=0$ allows the model to ignore information for hand $h$ with respect to sign $c$, and is equivalent to setting the probability for that hand to 1.

We can calculate the parameters $a_c^h$ in various ways;  the simplest approach, used in this paper, is to set these parameters according with the annotations provided for the dataset specifying which hands are employed for each class of signs.



\subsubsection{Exploiting the absence of a hand in the video}

When a hand $h$ is missing from a sample video, we can assume that there is no possibility that the sign in the video uses that hand. If from the annotations mentioned in the previous subsection we know that hand $h$ is used in the signs of class $c$ (i.e., $a_c^h=1$), but that hand is missing from the video, then the probability for a class $c$ in that case should be set to $0$.

We can compute $x_{a}^h$, a boolean variable with value $1$ when hand $h$ is present in a sign when testing. We consider a hand as present when it can be detected in more than $50\%$ of the frames of the video. With it we can define:

\begin{equation}
  P(x_{a}^h|c)=\begin{cases}
  0 & \mbox{if }  a_c^h=1 \; \text{and} \; x_a^h=0 \\
  1 & \mbox{otherwise}
  \end{cases}
\end{equation}

We can then add two factors (one for each hand) to Equation \ref{eq:model2} to penalize this situation, such that:


\begin{equation}
  \label{eq:model3}
  P(x|c)=P(x^l|c)^{a_c^l} P(x_{a}^l|c)  P(x^r|c)^{a_c^r} P(x_{a}^r|c)
\end{equation}

\section{Results}

\label{sec:results}

We performed experiments with the proposed model, setting parameters as described in the previous sections. We employed a stratified random subsampling cross-validation methodology, with an 80/20 training/testing split and 30 independent runs for each experiment. In the following subsections we present and analyze the results of the experiments.

\subsection{Subject-dependent experiments}
\label{section:general_experiments}

Table \ref{tab:dependent_results} shows the result of the subject-dependent experiments, where the model obtains an accuracy of 96.2\%. The table also shows the accuracies obtained by the model when using a subset of the features, to measure how much each information each of the corresponding subclassifiers is providing. All subclassifiers seem to provide non-redundant information, since the decreases in accuracy after removing a feature are all significant. Nonetheless, the position subclassifier can classify correctly a large amount (76.1\%) of signs by itself, which could be an effect due to the distribution of positions in the dataset.

The ALL-HMM column also shows the mean accuracy when replacing the trajectory and handshape subclassifiers with Hidden Markov Models with Gaussian Mixture Models output probabilities (HMM-GMM). We trained a model for each class and feature, using EM. Each model is a left-to-right HMM with skip transitions and 4 states in all cases. The handshape subclassifier also uses as input the output probabilities of the static handshape classifier described in \cite{Ronchetti2016}.

We also performed subject-dependent experiments with the binary features described in \cite{Kadir2004}, that include the same kind of information our classifier considers (position, movement and handshape). The features were calculated for each sample, and in each case the resulting $features \times frames$ matrix was resampled so that the number of frames was the same for all samples (32 frames in our experiments). The resulting $features \times 32$ matrices were used as input to a one-versus-all multiclass SVM with a linear kernel. The results of these experiments, that followed the same cross-validation scheme, are shown in column ALL-BF-SVM.


\begin{table}
\centering
\caption{Subject-dependent experiments, with various combinations of features. ALL: all features. HS: Handshape features. MOV: Movement features. POS: Position features. ALL-HMM: All features with HMM subclassifiers. ALL-BF-SVM: All features with Binary Features and a SVM classifier.}
\begin{tabular}{p{7mm}gfffffkkk}
  \toprule
  & ALL & HS & MOV & POS & HS-POS& HS-MOV & POS-MOV & ALL-HMM & ALL-BF-SVM\\
  \midrule
   $\mu$  & 97.44  & 52.97 & 54.03  & 76.05  & 94.91 & 83.59 & 84.84 & 95.92 & 95.08
   \\
   $\sigma$   & 0.59 &  1.74 & 1.71 & 0.62 & 0.52 & 0.87 & 0.90 & 0.95 & 0.69
  \\\addlinespace
 \bottomrule
\end{tabular}
\label{tab:dependent_results}
\end{table}


%

To test the confusion of the model between one and two handed signs, we performed independent experiments dividing the dataset in two subsets: one with the one-handed signs (1H, 42 classes) and another with the two-handed ones (2H, 22 classes) (Table \ref{tab:hands_experiments}). Since the weighted mean between the accuracies of this two experiments does not differ significantly from the subject-dependent mean from Table \ref{tab:dependent_results} (97.44\%), this provides evidence that the model is discriminating very well between one and two-handed signs.

\begin{table}
	\centering
	\caption{Accuracy of the model on the one-handed (1H) and two-handed (2H) subsets of the LSA64 dataset. The last column shows the mean accuracy between 1H and 2H, weighted by the percentage of signs in each subset.}
	\begin{tabular}{p{8mm}fff}
		\toprule
		  Subset  	&  1H & 2H  & Mean
		\\ \midrule
		$\mu$       & 95.93 & 99.09 & 97.01
		\\
		$\sigma$    & 1.31  &  0.77 & -
	\end{tabular}
	\label{tab:hands_experiments}
\end{table}

\begin{figure}
\minipage{0.49\textwidth}
	\centering
	\includegraphics[width=1.1\textwidth, height= 1.0\textwidth]{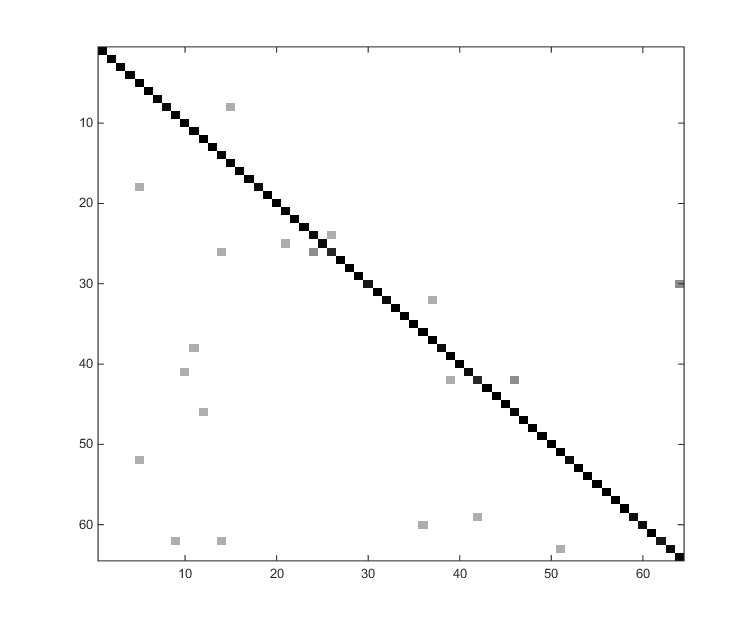}
	\caption{Confussion Matrix of the LSA64 database.}
	\label{fig:conf_matrix_LSA64}
\endminipage\hfill
\minipage{0.49\textwidth}
	\centering
	\includegraphics[width=1.1\textwidth, height= 1.0\textwidth]{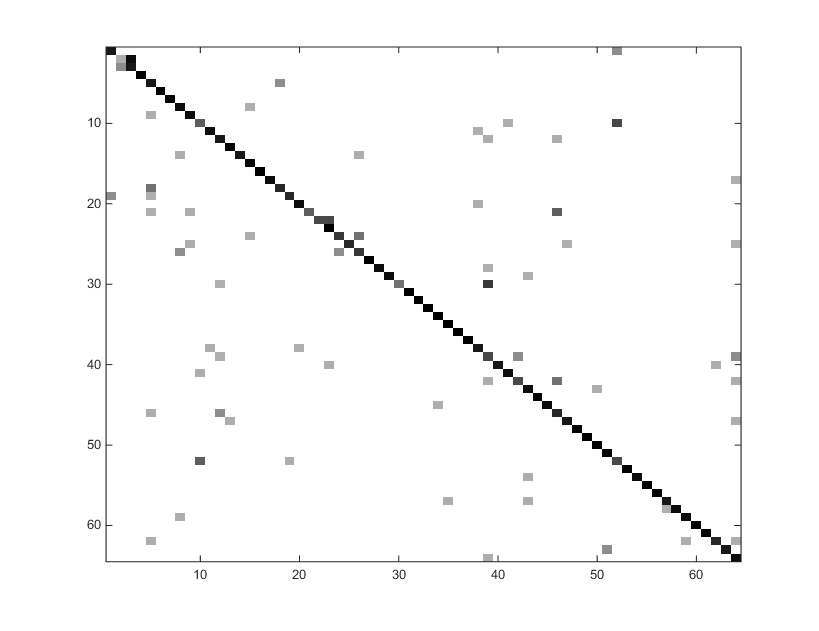}
	\caption{Confussion Matrix of the LSA64 without HandShapes Subclassifier.}
	\label{fig:conf_matrix_LSA64_noHS}
\endminipage
\end{figure}

Figure \ref{fig:conf_matrix_LSA64} shows the confusion matrix for all signs of the dataset. There are few visible patterns in the matrix, providing some evidence that in general the model does not suffer from biases. There are however some exceptions. For example, the model confuses classes 24 and 26, since both posses similar movements and positions, and have only a subtle difference in handshape. Figure \ref{fig:conf_matrix_LSA64_noHS} shows the same confusion matrix after removing the handshape classifier from the model. Here the confusion between classes 24 and 26 increases dramatically, since without  handshape information it is impossible to distinguish the two.

\subsection{Subject-independent experiments}

To evaluate how well the model generalizes with an unseen subject, we performed subject-independent experiments, where we trained the model with nine of the ten subjects and tested it with the remaining one. For each subject left out, we performed 30 runs of the experiment, whose results are show in Table \ref{tab:cross-subject-val}. As expected, the accuracy decreases on these experiments, but not overmuch.



\begin{table}
\centering
\caption{Subject-independent experiments on the LSA64 database. Each column shows the mean accuracy when testing with a subject. The final column shows total mean.}
\begin{tabular}{p{9mm}jjjjjjjjjjg}
	\toprule
     Subject   	& 1 & 2 & 3 & 4 & 5 & 6 & 7 & 8 & 9 & 10 & Mean
	\\ \midrule
	$\mu$      & 94.5 & 93.8 & 87.7 & 93.8 & 91.8 & 92.6 & 89.1 & 90.3 & 88.4 & 94.6 & 91.7
	\\
	$\sigma$   & 0.66  & 0.83  &  1.05 &  0.79 &  0.65 &  0.41  &  0.91 &  0.70 &  0.85 &  0.66 & 0.75
\end{tabular}
\label{tab:cross-subject-val}
\end{table}
\vspace{-20px}

\section{Conclusion}
We have presented a sign language recognition model that does not employ frame-sequence information and still achieves low classification error for both subject-dependent and independent tasks. We tested the model on a medium-sized Argentinian Sign Language dataset. We also compared the model to a sequence dependent one by replacing the trajectory and handshape subclassifiers with HMM-GMM models, and found little difference in the accuracy of both models, providing further evidence of the validity of the sequence-agnostic model. This approach could point the way to new ways of defining sub-units for sign language recognition. The model could also provide advantages for real-time recognition, specially for dealing with out-of-order or missing frames.

In future work, we plan on testing the model on continuous sign language tasks with a sentence-level dataset, as well as determining its suitability for real time tasks. We also plan on introducing new sub-classifiers to improve the model's performance.

\bibliographystyle{splncs03}
\bibliography{bib/general,bib/psom,bib/surrey,bib/triesch,bib/midusi}

\end{document}